%% file: template.tex
\documentclass{article}
\usepackage{amsmath,epsfig}
\usepackage[preprint]{spconfa4}
\input{math_commands.tex}

\usepackage{hyperref}
\usepackage{url}
\usepackage{graphics,float}
\usepackage{wrapfig,lipsum,booktabs}
\usepackage{xcolor}
\usepackage{subfigure}
\usepackage[]{caption}
\usepackage[utf8]{inputenc} 
\usepackage[T1]{fontenc}    
\usepackage{booktabs}       
\usepackage{amsfonts}       
\usepackage{nicefrac}       
\usepackage{microtype}      
\usepackage{color}
\usepackage{graphicx}
\usepackage{multirow}
\usepackage{multicol}
\usepackage[misc]{ifsym} 

\usepackage{algorithm}
\usepackage{algorithmic}
\usepackage{xspace}
\usepackage{multirow}
\usepackage{multicol}
\usepackage{subfigure}
\usepackage{epsfig,graphics,float}

\let\OLDthebibliography\thebibliography
\renewcommand\thebibliography[1]{
  \OLDthebibliography{#1}
  \setlength{\parskip}{0pt}
  \setlength{\itemsep}{0pt plus 0.3ex}
}

\begin{document}\sloppy

\def\x{{\mathbf x}}
\def\L{{\cal L}}

\title{Automated Graph Learning via Population Based Self-Tuning GCN}
%
\name{Ronghang Zhu$^{\ast}$, Zhiqiang Tao$^{\dagger}$, Yaliang Li$^{\ddagger}$, Sheng Li$^{\ast}$$^{(\textrm{\Letter})}$}
\address{$^{\ast}$Department of Computer Science, University of Georgia, Athens, GA\\
$^{\dagger}$Department of Computer Science and Engineering, Santa Clara University, Santa Clara, CA\\ 
$^{\ddagger}$Alibaba Group, Bellevue, WA\\
$\{$ronghangzhu, sheng.li$\}$@uga.edu, ztao@scu.edu, yaliang.li@alibaba-inc.com}

\maketitle

\begin{abstract}
Owing to the remarkable capability of extracting effective graph embeddings, graph convolutional network (GCN) and its variants have been successfully applied to a broad range of tasks, such as node classification, link prediction, and graph classification. Traditional GCN models suffer from the issues of overfitting and oversmoothing, while some recent techniques like DropEdge could alleviate these issues and thus enable the development of deep GCN. However, training GCN models is non-trivial, as it is sensitive to the choice of hyperparameters such as dropout rate and learning weight decay, especially for deep GCN models. In this paper, we aim to automate the training of GCN models through hyperparameter optimization. To be specific, we propose a self-tuning GCN approach with an alternate training algorithm, and further extend our approach by incorporating the population based training scheme. Experimental results on three benchmark datasets demonstrate the effectiveness of our approaches on optimizing multi-layer GCN, compared with several representative baselines.     
\end{abstract}
\section{Introduction}
Graph-structured data are ubiquitous in various real-world applications, which promotes the demand of expanding deep learning techniques to graphs~\cite{zhou2018graph}. Many approaches have been proposed to learn node feature representations by investigating graph convolutional networks (GCN)~\cite{GCN,GAT,GraphSAGE,jiang2020co}. However, these GCN models are usually no deeper than three or four layers. Recently, several attempts have been proposed to explore deeper GCN models, such as incorporating residual/dense connections and dilated convolutions to build deeper GCN models~\cite{li2019deepgcns, chen2020simple}, and adopting the idea of  DropEdge~\cite{dropEdge} to solve overfitting and oversmoothing problems in deeper GCN. Most of these approaches focus on designing an appropriate deeper GCN structure yet ignoring the importance of hyperparameter choices to train deeper GCN models.

Recently, automatically optimizing hyperparameters has yielded remarkable results on many machine learning tasks. There are mainly two categories: One is hyperparameter configuration search, like random search~\cite{RandomSearch}, grid search~\cite{GridSearch} and Hyperband~\cite{Hyperband}, which optimizes hyperparameters in fixed values during training process. The other is hyperparameter schedule search such as self-tuning networks (STN)~\cite{STN} and population based training (PBT)~\cite{PBT}, which enable hyperparameters to change in each training iteration. To the best of our knowledge, the hyperparameter optimization problem for graph neural networks has not been studied yet.

In this paper, we propose a novel automated graph learning algorithm to investigate deeper GCN models. Different from existing works on graph neural architecture search~\cite{ijcai2020-195,Auto-GNN,Policy-GNN}, our work focuses on automatically tuning hyperparameters with given GCN architectures. The major contributions of this paper are summarized as follows.
\begin{itemize}
    \item We propose to solve the automated graph learning problem from a new perspective, \emph{i.e.}, through hyperparameter optimization, which provides a complementary direction to graph neural architecture search.  
    \item We design and develop self-tuning GCN (ST-GCN) by incorporating hypernets~\cite{STN} in each graph convolutional layer, enabling a joint optimization over GCN model parameters and its hyperparameters. The proposed approach can be flexibly extended to many existing GCN models~\cite{jiang2019censnet,GAT,GraphSAGE}.
    \item We further adopt a population based training framework to self-tuning GCN, which alleviates local minima problem by exploring hyperparameter space globally.
    \item We conduct extensive experiments to demonstrate the effectiveness of the proposed approaches on three benchmark datasets.
\end{itemize}

\section{Related Work}
\textbf{Graph Neural Networks} (GNNs) have been a mainstream technique to squeeze complex graph-structured data into compact and low-dimensional embedding representations~\cite{zhou2018graph,jiang2019censnet,saed2021edge,zhu2021transferable,sheu2020context}. Roughly, it could be separated into two categories: spectral-based GNNs~\cite{SpectalCNN,GCN} and spatial-based GNNs~\cite{GAT,GraphSAGE}. The former develops graph convolution operations in the vein of spectral graph theory; the latter instantiates convolution on spatial domain, relying more on neighborhood sampling, message passing, and feature aggregation. To name a few, graph convolutional networks (GCN)~\cite{GCN} and graph attention networks (GAT)~\cite{GAT} are two representative spectral and spatial-based methods, respectively, which are widely used as building modules in other graph learning frameworks. In this paper, we focus our study on the first category, with a particular interest in the impact of hyperparameters on GCN.

\textbf{Hyperparameter Optimization} (HPO)~\cite{Feurer2019} lies in the core task of AutoML, which aims to optimize the model performance by automatically searching feasible hyperparameter configurations or schedules. Some representative HPO methods, yet not limited to, include grid/random search~\cite{RandomSearch}, Hyperband and successive halving~\cite{Hyperband}, hypergradient based method~\cite{pmlr-v70-franceschi17a}, and Bayesian optimization~\cite{BOHB}. Recent research~\cite{dropEdge} has shown that different hyperparameters, such as learning rates, dropout, weight decay, etc., largely impact the model performance of various GNN architectures. Thus, it is a promising direction to incorporate HPO into GNNs to enable automated learning. In light of this, we develop a Self-tuning GCN (ST-GCN) model with population-based training, inspired by the recent hyperparameter scheduling methods~\cite{PBT,STN,HPM}. To our best knowledge, this is the first attempt to propose a GNN-specific hyperparameter optimization algorithm.

\textbf{Automated Graph Learning} has emerged as an important research problem that combines AutoML and GNNs. Similar to neural architecture search (NAS), existing automated graph learning methods mainly target to explore an optimal GNN architecture from a pre-defined network configuration space~\cite{ijcai2020-195,Auto-GNN}. The searching space is generally defined by GNN structures, including network depth, aggregation and activation functions, etc., and the search processing is governed by a controller model to optimize the performance on the validation set. Following this line, a series of graph neural architecture search (GNAS) methods have been proposed recently, implemented by reinforcement learning~\cite{ijcai2020-195,Auto-GNN,Policy-GNN} and evolution algorithm~\cite{shi2020evolutionary,jiang2020graph}. Unlike GNAS, the proposed ST-GCN studies ``automation'' from a new perspective, \emph{i.e.}, hyperparameter optimization, which automatically tunes hyperparameters for pre-defined network architecture, and thus serves as a complementary direction towards automated graph learning.

\section{Methodology}
\subsection{Preliminary}

Let $\mathcal{G} = (\mathcal{V}, \mathcal{E})$ denote an undirected graph with $N$ nodes $v_i \in \mathcal{V}\; (i=1,\cdots,N)$ and a number of edges $(v_i, v_j)\in \mathcal{E}$. The adjacency matrix of graph $\mathcal{G}$ is denoted by $A \in \mathbb{R}^{N\times N}$, where $A_{ij}=1$ if there is an edge between $v_i$ and $v_j$. We consider the node classification task~\cite{NodeClassify} on graph $\mathcal{G}$, and use $\mathcal{Y}$ with $y_i \in \mathbb{R}\; (i=1,\cdots,N)$ to denote the node labels. Moreover, we define a graph learning model (e.g., GCN) as $f(\cdot;\theta,\lambda):\mathcal{V}\rightarrow{\mathcal{Y}}$, which is parameterized by the model parameters $\theta \in \mathbb{R}^p$ (e.g., weights ans biases) and hyperparameters $\lambda \in \mathbb{R}^q$ (e.g., dropout rate and weight decay).

For node classification, the graph learning model $f(\cdot; \theta, \lambda)$ can be optimized by solving:
\begin{equation}
\label{objective}
\mathcal{L}(\theta,\lambda)=\mathbb{E}_{(v,y)\in\mathcal{D}}\big[\ell(f(v;\theta,\lambda),y) \big],
\end{equation}
where $\ell(\cdot,\cdot)$ refers to a loss function and $\mathcal{D}$ represents a training set $\mathcal{D}_{trn}$ or a validation set $\mathcal{D}_{val}$. Upon Eq.~\eqref{objective}, we can see the loss value depends on both the model parameters $\theta$ and the hyperparameters $\lambda$. Traditionally, the selection of  hyperparameters $\lambda$ is an iterative manner with trial and error required profound knowledge of machine learning algorithms and statistics. In this paper, we aim to design an approach to automatically choose optimal hyperparameters for graph learning models.

The model $f(v;\theta,\lambda)$ in Eq.~\eqref{objective} could be implemented by various graph learning approaches, such as the traditional graph embedding methods and the recent graph neural networks. In this paper, we focus on graph convolutional networks (GCN)~\cite{GCN}. However, it is worth noting that, the proposed hyperparameter optimization method can be easily adapted to other graph neural networks. Given an input graph characterized by a normalized adjacency matrix $\hat{A}$ and a node feature matrix $H^{(0)}=X$, GCN updates the node embeddings by using the following layer-wise propagation rule: 
\begin{equation}
\label{GCN}
H^{(l+1)} = \sigma\Big(\hat{A}H^{(l)}W^{(l)} \Big),
\end{equation}
where $\hat{A}=\hat{D}^{-1/2}(A+I)\hat{D}^{-1/2}$ is the normalized adjacency matrix, $I$ is an identity matrix, and $\hat{D}$ is the degree matrix of $A+I$. $W^{(l)}\in\mathbb{R}^{M_l\times M_{l+1}}$ is the learnable weight matrix in the $l$-th layer with $M_l$ refers to the feature dimension of $H^{(l)}$. $H^{(l)}=\{h^{(l)}_1, h^{(l)}_2, \ldots, h^{(l)}_N\}$ is the hidden feature matrix with $h^{(l)}_i$ as the hidden feature for node $v_i$. $\sigma(\cdot)$ denotes an activation function such as the ReLu function.

Then, the graph learning model with two-layer GCN for node classification is defined as
\begin{equation}
\label{GCN2layer}
f(\mathcal{V};\theta,\lambda) = \text{softmax}\Big(\hat{A}\text{ReLu}\Big(\hat{A}\mathcal{V}W^{(0)}\Big)W^{(1)}\Big).
\end{equation}

As mentioned in~\cite{GCN}, overfitting is one of the main obstacles to build deep GCN model for node classification. To alleviate this issue, DropEdge~\cite{dropEdge} is proposed to randomly drop out a certain part of edges in the graph, which is defined as
\begin{equation}
\label{DropEdge}
A_{drop} = A - A'.
\end{equation}
Here $A'$ denotes a random subset of edges from original $A$, $A_{drop}$ refers to the resulting adjacency matrix after dropped edges. Replaced $\hat{A}$ with $\hat{A}_{drop}$ in deep GCN model for propagation can prevent overfitting problem, where $\hat{A}_{drop}$ is the re-normalized $A_{drop}$. In the following, we consider four-layer and eight-layer GCN models with DropEdge~\cite{dropEdge} for node classification on a graph.

\subsection{Self-Tuning GCN}

We propose a self-tuning GCN (ST-GCN) approach to guide the hyperparameter search of GCN models. In particular, we define $\hat{\theta}(\lambda):\mathbb{R}^q\rightarrow{\mathbb{R}^p}$ as the response function of hyperparameter $\lambda$ to approximate the GCN model parameter $\theta$, i.e.,  $\hat{\theta}(\lambda)$ is a mapping from $\lambda$ to optimal parameters. For a given GCN layer with the weight matrix $W\in\mathbb{R}^{M_{in}\times{M_{out}}}$ and bias $b\in\mathbb{R}^{M_{out}}$, we define the affine transformation of hyperparameters $\lambda$ as
\begin{equation}
\label{STLayer}
\hat{W}(\lambda) = W + W_{\lambda}\odot_{row} C_{W}(\lambda) \;\; \text{and} \;\; \hat{b}(\lambda) = b + b_{\lambda}\odot C_{b}(\lambda),
\end{equation}
where the dimensions of $W_{\lambda}$ and $b_{\lambda}$ are the same as $W$ and $b$, respectively. $C_{W}(\lambda)\in\mathbb{R}^{M_{out}}$ and $C_{b}(\lambda)\in\mathbb{R}^{M_{out}}$ are scaled embeddings by linearly transforming $\lambda$, i.e., $C_{w}(\lambda)=e_{w}\lambda$ and $C_{b}(\lambda)=e_{b}\lambda$. $\odot_{row}$ denotes the row-wise rescaling and $\odot$ indicates element-wise multiplication. Thus the total parameters of the GCN model are $\hat{\theta}(\lambda)=\{\hat{W}(\lambda), \hat{b}(\lambda)\}$.

As $\hat{\theta}(\lambda)$ captures changes such as shifting and scaling in $\theta$ induced by $\lambda$, we reformulate Eq.~\eqref{objective} with $\hat{\theta}(\lambda)$ as
\begin{equation}
\label{objectiveTrain} 
\mathcal{L}(\hat{\theta}(\lambda), \lambda) = \mathbb{E}_{\lambda \sim P(\lambda),(v,y)\in \mathcal{D}}\Big[\ell(f(v;\hat{\theta}(\lambda), \lambda), y) \Big].
\end{equation}
Here, $P(\lambda)=P(\lambda|\epsilon)$ represents a log-uniform distribution over hyperparameter $\lambda$. $\epsilon$ control the sccale of the hyperparameter distribution, which contains the bounds of the ranges of $\lambda$. $\hat{\theta}(\lambda)$ can capture the best-response over the samples and the shape locally around the hyperparameter values. We vary the distribution $P(\lambda|\epsilon)$ with training iterations. To prevent $P(\lambda|\epsilon)$ from collapsing to a degenerate distribution, we add an entropy regularization item $\mathbb{H}[\cdot]$ weighted by $\tau \in \mathbb{R}_{+}$. The objective function in~Eq.~\eqref{objectiveTrain} becomes:
\begin{small}
\begin{equation}\label{objectHyper}
\mathcal{L}(\hat{\theta}(\lambda), \lambda) =\mathbb{E}_{\lambda \sim P(\lambda|\epsilon),(v,y)\in \mathcal{D}}\Big[\ell(f(v;\hat{\theta}(\lambda), \lambda), y) \Big] - \tau\mathbb{H}[P(\lambda|\epsilon)].
\end{equation}
\end{small}

\subsection{Alternate Training Algorithm}
To optimize the objective functions of ST-GCN in Eq.~\eqref{objectiveTrain} and Eq.~\eqref{objectHyper}, we follow the alternate training procedure proposed by~\cite{STN}, which includes two steps, i.e., $ModelTraining$ and $HyperTraining$. In particular, we use $\mathcal{L}_{trn}$ to denote the $ModelTraining$ loss on training set $\mathcal{D}_{trn}$ following Eq.~\eqref{objectiveTrain}, and use $\mathcal{L}_{val}$ to denote the $HyperTraining$ loss on validation set $\mathcal{D}_{val}$ following Eq.~\eqref{objectHyper}.

In the $ModelTraining$ step, we adopt a stochastic gradient update of $\theta$ to minimize Eq.~\eqref{objectiveTrain} with sampling $\lambda\sim P(\lambda|\epsilon)$. Specifically, $\theta$ is updated by
\begin{equation}
\label{thetaUpdate}
\theta_{(t)} \leftarrow \hat{\theta}_{(t-1)}(\lambda_{(t-1)}) - \eta_{\theta}\nabla\theta,
\end{equation}
where $\eta_{\theta}$ is the learning rate, $\nabla\theta = \frac{\partial{\mathcal{L}_{trn}}}{\partial{\theta}}$ is the gradient of model parameter.

In the $HyperTraining$ step, we make a stochastic gradient update of $\lambda$ and $\epsilon$ to minimize Eq.~\eqref{objectHyper}. In detail, $\lambda$ is updated by
\begin{equation}
\label{lambdaUpdate}
\lambda_{(t)} \leftarrow \hat{\lambda}_{(t-1)} - \eta_{\lambda}\nabla\lambda,
\end{equation}
where $\eta_{\lambda}$ is the learning rate, and $\nabla\lambda$ is the hypergradient given by
\begin{equation}
\label{lambdaUpdate}
\nabla\lambda = \frac{\partial{\mathcal{L}_{val}(\hat{\theta}(\lambda), \lambda)}}{\partial{\theta}}\frac{\partial{\theta}}{\partial{\lambda}} + \frac{\partial{\mathcal{L}_{val}(\hat{\theta}(\lambda), \lambda)}}{\partial{\lambda}}.
\end{equation}

The computation in Eq.~\eqref{lambdaUpdate} is mainly determined by response function $\hat{\theta}(\lambda)$, which is memory-efficient and flexible to compute~\cite{STN}. We summarize the procedures of ST-GCN in Algorithm~\ref{alg:STG}.

\begin{algorithm}[t]
\caption{Self-Tuning GCN}
\label{alg:STG}
\begin{algorithmic}[1]
\REQUIRE 
Hyperparameter $\lambda$, GCN model parameter $\theta$, distribution scale $\epsilon$ response function $\hat{\theta}(\cdot)$, learning rate $\eta_{\lambda}$, $\eta_{\theta}$ and $\eta_{\epsilon}$.
\ENSURE
GCN model parameter $\theta$, hyperparameter $\lambda$
\WHILE{not converged }
\FOR{$i = 1,\ldots,T_{trn}$}
\STATE $\theta \leftarrow \theta - \eta_{\theta}\frac{\partial \mathcal{L}_{trn}}{\partial \theta}$
\ENDFOR
\FOR{$i = 1,\ldots,T_{val}$}
\STATE $\lambda \leftarrow \lambda - \eta_{\lambda}\frac{\partial \mathcal{L}_{val}}{\partial \lambda}$, $\epsilon \leftarrow \epsilon - \eta_{\epsilon}\frac{\partial \mathcal{L}_{val}}{\partial \epsilon}$, $\epsilon \in P(\lambda|\epsilon)$
\ENDFOR
\ENDWHILE
\RETURN $\theta$, $\lambda$
\end{algorithmic}
\end{algorithm}

\begin{algorithm}[t]
\caption{Population based Self-Tuning GCN}
\label{alg:PSTG}
\begin{algorithmic}[1]
\REQUIRE An initialized population of GCN models $\mathcal{S}=\big\{s^k\big\}_{k=1}^K$, where $s^k = f(\cdot;\theta^k, \lambda^k)$.
\ENSURE
GCN model parameter $\theta$, hyperparameter $\lambda$
\FOR{$(\theta, \lambda, s, t)\in \mathcal{S}$ (asynchronously in parallel)}
    \WHILE{not end of training step}
        \STATE $\theta \leftarrow ModelTraining(\mathcal{D}_{trn};\theta, \lambda)$
        \STATE $(\lambda, \epsilon) \leftarrow HyperTraining(\mathcal{D}_{val};\theta, \lambda)$
        \STATE $acc_s \leftarrow eval(\mathcal{D}_{val}, s)$
    \IF{$ready(s, t, \mathcal{S})$}
        \STATE $(\theta',\lambda') \leftarrow Exploitation(\theta,\lambda,s, \mathcal{S})$
        \IF{$\theta \neq \theta'$}
            \STATE $(\theta, \lambda) \leftarrow Exploration(\theta', \lambda')$, $acc_s \leftarrow eval(\mathcal{D}_{val}, s)$
        \ENDIF
    \ENDIF
    \STATE update $\mathcal{S}$ with new $(\theta, \lambda, s, t+1)$
    \ENDWHILE
\ENDFOR
\RETURN $\theta$, $\lambda$ with the highest $acc_s$ of $s$ in $\mathcal{S}$
\end{algorithmic}
\end{algorithm}

\subsection{Population based Self-Tuning GCN}
Our ST-GCN approach mainly focuses on computing hypergradients by designing a differentiable response function for hyperparameter, which is a non-convex optimization problem and may get into local minima~\cite{HPM}. To address this potential issue with ST-GCN, we propose a population based self-tuning GCN (PST-GCN) approach, inspired by the recent success of population based training~\cite{PBT,HPM}. The population based training (PBT) could supply abundant multiplicity to globally seek hyperparameters throughout the hyperparameter space, which employs a population of agent models to search different hyperparameter combinations and update hyperparameters with a mutation operation. PBT is a good complementary to ST-GCN, as it could help overcome the local minima issue.

Given a population of agent models $\mathcal{S}_{(t)}=\big\{s_{(t)}^k\big\}_{k=1}^{K}$ at the $t$-th training step, where $s_{(t)}^k$ denotes the $k$-th agent model \text{w.r.t} $f(\cdot;\theta,\lambda)$, and $K$ refers to the population size. The searching process of Population based Self-Tuning GCN (PST-GCN) includes three types of operations as follows:
\begin{itemize}
\itemsep0em    
\item  \textbf{Training step} evaluates the accuracy on the validation dataset,  $eval\big(\mathcal{D}_{val}, s_{(t)}^k\big)$,  and updates $\theta_{(t-1)}^k$ to $\theta_{(t)}^k$, $\lambda_{(t-1)}^k$ to $\lambda_{(t)}^k$ and $\epsilon_{(t-1)}^k$ to $\epsilon_{(t)}^k$. The training step has a fixed number of epochs. After the training step, the agent $s_{(t)}^k$ is ready for exploitation and exploration.
\item \textbf{Exploitation} operation exploits the population of agent models $\mathcal{S}_{(t)}$ by dividing it into three subsets of $top$, $middle$ and $bottom$ in terms of the accuracy of validation. The parameters and hyperparameters in the $bottom$ agent models are replaced by the $top$ ones.
\item \textbf{Exploration} operation maintains the $top$ and $middle$ agents unchanged, and randomly perturb the hyperparameters of $bottom$ agents. 
\end{itemize}

With the above three operations, the proposed PST-GCN can have a balance between local and global search to overcome the potential local minima issue in ST-GCN. Algorithm~\ref{alg:PSTG} summarizes the main procedures of our PST-GCN approach.

\section{Experiments}

\textbf{Datasets}. We employ three benchmark citation network datasets for experiments, including Cora, Citeseer and Pubmed~\cite{datasets}. Each document contains a sparse bag-of-words feature vector, and there is a list of citation links between documents. We treat the documents as nodes and citation links as undirected edges. The statistics of these three datasets are summarized in Table~\ref{benchmark}. Label rate denotes the percentage of labeled nodes among all nodes for model training. We follow the settings in ~\cite{GCN} and use the full-supervised training fashion on all datasets in experiments.

\begin{table}[htb]
\centering
\caption{Statistics of three benchmark graph datasets.}
\vspace{-6pt}
\resizebox{\textwidth/2}{!}{
\tiny\begin{tabular}{lccccc}
\toprule
Dataset  &Nodes  &Edges &Classes &Features &Label Rate \\
\midrule
Cora  &2,708 &5,429 &7 &1,433 &0.052 \\
Citeseer &3,327 &4,732 &6 &3,703 &0.036 \\
Pubmed &19,717 &44,338 &3 &500 &0.003 \\ 
\bottomrule
\end{tabular}}
\label{benchmark}
\end{table}

\noindent \textbf{Experiment Setting}. We adopt two different backbones of GCN with 4 layers and 8 layers separately. The hyperparameters considered in our experiments include variational dropout rates for hidden state, edge dropout rate~\cite{dropEdge} and weight decay. The numbers of hyperparameters are 5 and 9 for the 4-layer and 8-layer GCN model, respectively.

\noindent \textbf{Baselines}. We compare the proposed methods ST-GCN and PST-GCN with three representative baselines including: (1) random search (RS): the best single model after 200 trials; (2) Hyperband (HB)~\cite{Hyperband}: the best single model after 200 trials; and (3) PBT~\cite{PBT}: the best single model in 200 agent models.



\noindent \textbf{Implementation Details}. We train the 4-layer and 8-layer GCN models with 128 hidden units per layer with all the above methods. Each model is trained for a maximum of 400 epochs. For PBT and PST-GCN, we set 200 epochs as warm up and 200 epochs as training iterations. After warm-up, we take one training epoch as one training step, and then perform exploitation and exploration after each training step. For ST-GCN and PST-GCN, we use an alternating training schedule to update the model parameters for two epochs on the $\mathcal{D}_{trn}$ and then update the hyperparameters for one epoch on the $\mathcal{D}_{val}$. We adopt the Adam optimizer for model training. For fixed hyperparameter baselines, i.e., RS and HB, the learning rate is set to 0.01 on the Cora and Pubmed datasets, and 0.09 on the Citeseer dataset, which follows the same settings in~\cite{dropEdge}. For the hyperparameter schedule baseline PBT and our approaches, we set the learning rate to 0.0005 on the Cora and Citeseer datasets, and 0.005 on the Pubmed dataset. The search spaces for the hyperparameters are as follows: dropout rates are in the range $[0,\;0.9]$, edge dropout rates~\cite{dropEdge} are in the range $[0,\;0.9]$, and weight decay is in the range $[10^{-6},\;10^{-2}]$.


\begin{table*}[t]
\centering
\caption{Node classification accuracy (in percentage) of our approaches (ST-GCN and PST-GCN) and three baselines (RS, HB and PBT) on Core, Citeseer and Pubmed datasets.}
\vspace{-6pt}
\begin{tabular}{ccccccc}
\toprule
\multirow{2}*{Dataset} &\multirow{2}*{Layer} &\multicolumn{5}{c}{Method}\\
\cmidrule(lr){3-7}
& & RS & HB & PBT & ST-GCN & PST-GCN \\
\midrule
\multirow{2}*{\shortstack{Cora}} &4 &85.3 &84.5 &82.9 &86.2 &\textbf{86.9} \\
&8 &83.5 &82.9 &84.4 &85.6 &\textbf{87.0} \\
\midrule
\multirow{2}*{\shortstack{Citeseer}} &4 &76.3 &76.5 &74.3 &\textbf{79.3} &79.0 \\
&8 &74.9 &74.8 &73.8 &78.1 &\textbf{78.6} \\
\midrule
\multirow{2}*{\shortstack{Pubmed}} &4 &\textbf{90.0} &89.3 &88.4 &89.6 &89.8 \\
&8 &87.4 &86.5 &88.2 &89.4 &\textbf{90.4} \\
\bottomrule
\end{tabular}
\label{results}
\end{table*}

\begin{figure}[htb]
\centering
\subfigure[Accuracy]{
\begin{minipage}[t]{0.5\linewidth}
\centering
\includegraphics[width=1.7in]{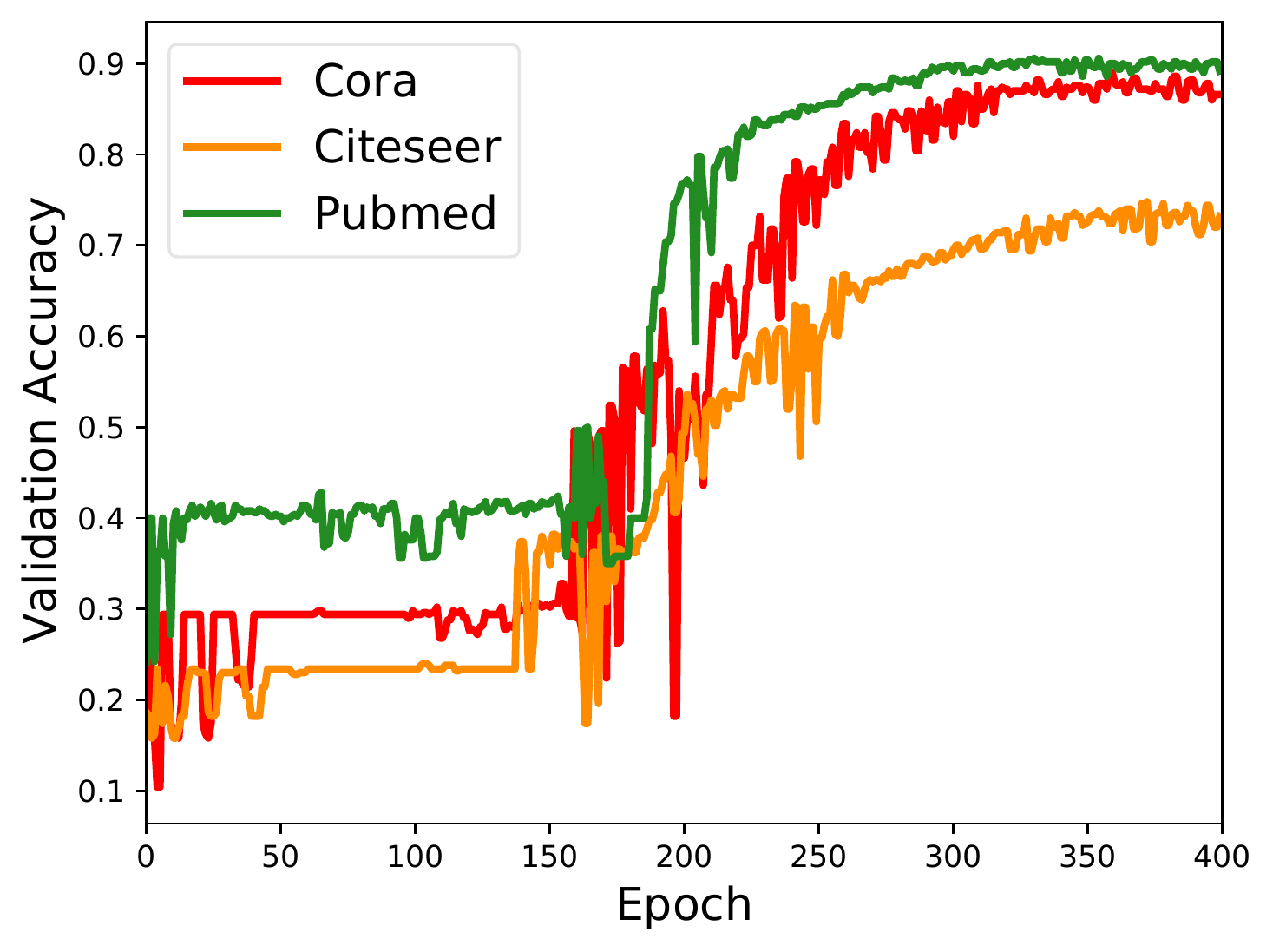}
\vspace{-12pt}
\label{val_acc}
\end{minipage}%
}%
\subfigure[Loss]{
\begin{minipage}[t]{0.5\linewidth}
\centering
\includegraphics[width=1.7in]{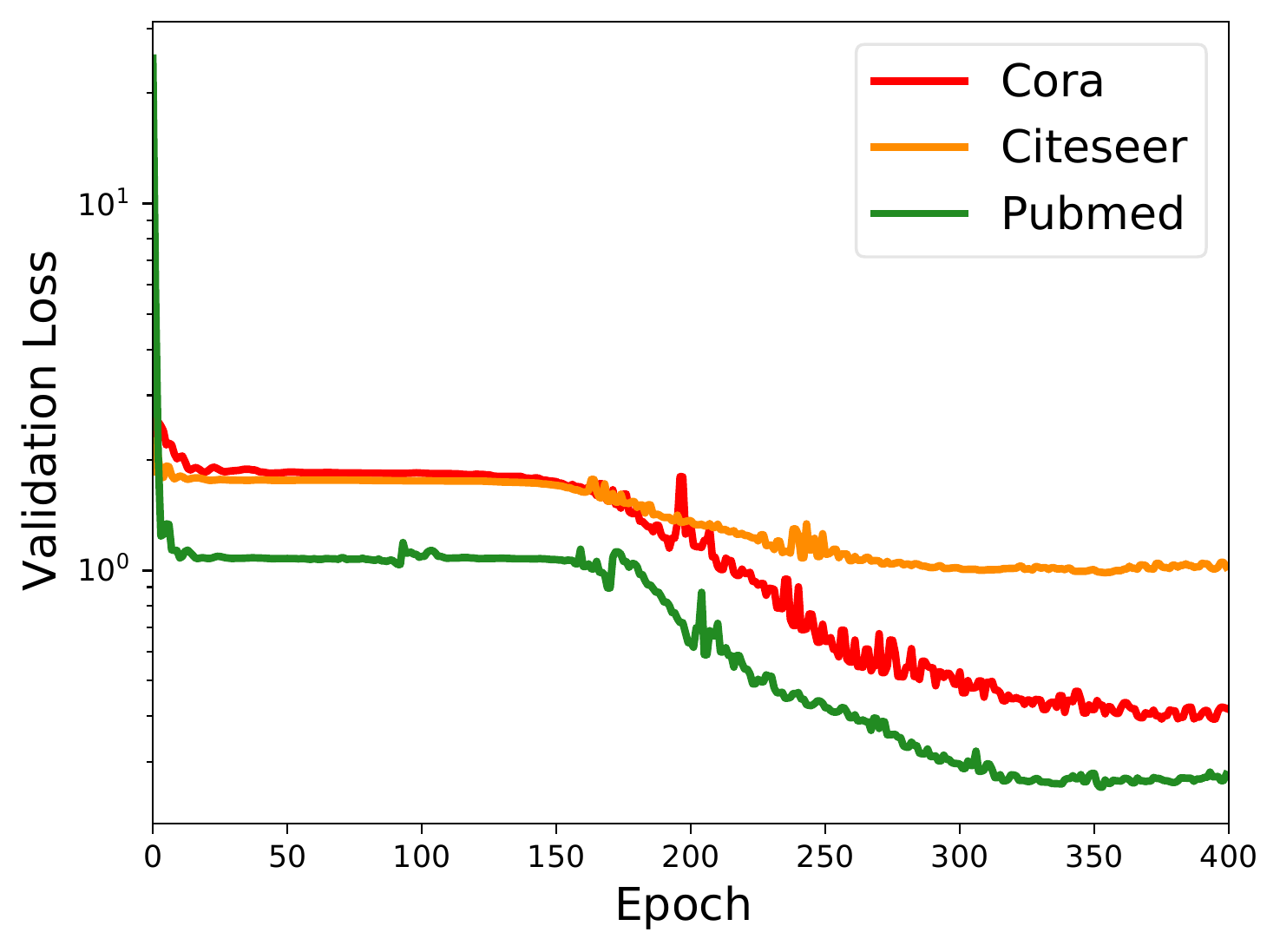}
\vspace{-12pt}
\label{val_loss}
\end{minipage}%
}%
\centering
\vspace{-12pt}
\caption{Experiments on Core, Citeseer and Pubmed dataset with 8-layer GCN. (a) The validation accuracy of ST-GCN over training epochs. (b) The validation loss of ST-GCN over training epochs.}
\vspace{-6pt}
\label{val_accloss}
\end{figure}

\noindent \textbf{Results and Analysis}. Performance metric is the classification accuracy on testing set in percent, and results are summarized in Table~\ref{results}. From it, we can observe that the proposed methods, ST-GCN and PST-GCN, obtain the best accuracy in 5 out of 6 experiments. Furthermore, compared with baselines on 4-layer and 8-layer GCN models, ST-GCN and PST-GCN have better advantages on 8-layer GCN models, which is due to the mighty power of gradient-based HPO in sophisticated hyperparameter space. Note that comparing with ST-GCN, PST-GCN can boost performance significantly on 8-layer GCN models, showing that population-based training can help alleviate the local minima problem in ST-GCN. 
Further, Fig.~\ref{val_acc} and Fig.~\ref{val_loss} show the validation accuracy and loss of ST-GCN over training epochs on 8-layer GCN model. We can observe that the validation accuracy is continuously improved along with the training epoch, while the validation loss curves are continuously decreased, even after a large number of epochs, which validates the effect of ST-GCN on alleviating overfitting in deeper GCN models.

\section{Conclusions}
In this paper, we study the automation of graph learning from the perspective of hyperparameter optimization, which is complementary to the existing GNN architecture search. To fulfill this goal, we propose a self-tuning GCN by jointly optimizing GCN model parameters and the hyperparameters, which can alleviate the overfitting in deeper GCN models. We further incorporate the population-based training to alleviate the local minima problem in ST-GCN, which provides a global hyperparameter search. Experimental results on three benchmark datasets demonstrate the benefit of the proposed population based self-tuning GCN method. 

\bibliographystyle{IEEEbib}
\small{\bibliography{template}}

\end{document}

%% file: math_commands.tex
\usepackage{amsmath,amsfonts,bm}

\def\eqref#1{equation~\ref{#1}}

\def\1{\bm{1}}

\DeclareMathAlphabet{\mathsfit}{\encodingdefault}{\sfdefault}{m}{sl}
\SetMathAlphabet{\mathsfit}{bold}{\encodingdefault}{\sfdefault}{bx}{n}

